\documentclass{article}

\usepackage[main, final]{neurips_2026}

\usepackage[utf8]{inputenc}
\usepackage[T1]{fontenc}
\usepackage{url}
\usepackage{booktabs}
\usepackage{amsfonts}
\usepackage{nicefrac}
\usepackage{microtype}
\usepackage{xcolor}
\usepackage{amsthm}

\usepackage[colorlinks=true,allcolors=blue]{hyperref}%
\hypersetup{
    colorlinks,
    linkcolor={blue!80!black},
    citecolor={blue!80!black},
    urlcolor={blue!80!black}
}
\usepackage{footnotebackref}

\usepackage{mathtools}
\usepackage{latexsym}
\usepackage{relsize}
\usepackage{tikz}
\usetikzlibrary{calc,positioning,fit,shapes.geometric}
\usepackage{subcaption}
\usepackage{multirow,array}
\usepackage{float}
\usepackage{color}
\usepackage{nicefrac}
\usepackage{comment}
\usepackage{booktabs}
\usepackage{natbib}
\usepackage{setspace}
\usepackage[nameinlink,capitalise]{cleveref}
\usepackage{makecell}
\usepackage{algorithm}
\usepackage{algorithmic}
\usepackage{braket}
\usepackage{amsfonts}
\usepackage{amssymb}
\usepackage[shortlabels]{enumitem}
\usepackage[bbgreekl]{mathbbol}
\usepackage{xcolor}
\usepackage{array}
 \usepackage{pgfplots}
\pgfplotsset{compat=1.11}
\usepgfplotslibrary{fillbetween}
\usetikzlibrary{intersections}

\newcolumntype{P}[1]{>{\centering\arraybackslash}p{#1}}
\usepackage{thm-restate}

\renewcommand{\mathbf}{\boldsymbol}

\DeclareCaptionLabelFormat{andtable}{#1~#2  \&  \tablename~\thetable}

\renewcommand{\mathbf}{\boldsymbol}
\newcommand{\bw}{{\boldsymbol{w}}}

\usetikzlibrary{fit}
\usetikzlibrary{shapes.geometric}

\DeclareSymbolFontAlphabet{\mathbb}{AMSb}
\DeclareSymbolFontAlphabet{\mathbbl}{bbold}

\def\X{{\mathcal X}}

\def\Y{{\mathcal Y}}

\def\A{{\mathcal A}}

\newcommand{\C}{\mathcal{C}}

\newcommand{\D}{\mathcal{D}}

\newcommand{\F}{\mathcal{F}}

\def\x{\mathbf{x}}

\def\w{\mathbf{w}}

\def\bz{\mathbf{z}}
\def\bp{\mathbf{p}}
\def\bq{\mathbf{q}}

\newcommand{\beps}{\boldsymbol{\epsilon}}

\DeclareMathAlphabet{\mathbfsf}{\encodingdefault}{\sfdefault}{bx}{n}

\DeclareMathOperator*{\conv}{conv}

\let\Pr\relax
\DeclareMathOperator{\Pr}{\mathbb{P}}

\newcommand{\lrset}[1]{\left\{#1\right\}}

\DeclarePairedDelimiter{\ceil}{\lceil}{\rceil}

\renewcommand{\O}{\mathcal{O}}

\newcommand{\ind}[1]{\mathbb{I}\!\lrset{#1}}

\newcommand{\eps}{\varepsilon}

\renewcommand{\leq}{~\le~}
\renewcommand{\geq}{~\ge~}

\let\oldtfrac\tfrac
\renewcommand{\tfrac}[2]{\smash{\oldtfrac{#1}{#2}}}

\let\nablaold\nabla
\renewcommand{\nabla}{\nablaold\mkern-2.5mu}

\usepackage{bbm}

\renewcommand{\eps}{\epsilon}

\def\beps{\boldsymbol{\epsilon}}

\DeclareSymbolFontAlphabet{\mathbb}{AMSb}

\newcommand{\RR}{\mathbb{R}}

\newcommand{\errs}{\ensuremath{\mathsf{errs}}}

\newcommand{\FP}{\mathrm{FP}}
\newcommand{\FN}{\mathrm{FN}}

 \newtheorem{theorem}{Theorem}
\newtheorem{lemma}[theorem]{Lemma}

 \newtheorem{definition}[theorem]{Definition}

\newtheorem{claim}[theorem]{Claim}

\title{Generalized Adaptive Boosting\\ and the Geometry of Mistakes}

\author{%
  Marco Bressan\\
  Università degli Studi di Milano\\
  \texttt{marco.bressan@unimi.it}\\
  \And
  Nataly Brukhim\\
  Apple Research\\
  \texttt{natalybrukhim@gmail.com}\\
  \And
  Nicolò Cesa-Bianchi\\
  Università degli Studi di Milano\\
  \texttt{nicolo.cesa-bianchi@unimi.it}\\
  \And
  Emmanuel Esposito\\
  Università degli Studi di Milano\\
  \texttt{emmanuel@emmanuelesposito.it}\\
  \And
  Yishay Mansour\\
  Tel Aviv University\\
  Google Research\\
  \texttt{mansour.yishay@gmail.com}\\
  \And
  Shay Moran\\
  Technion\\
  Google Research\\
  \texttt{smoran@technion.ac.il}\\
  \And
  Maximilian Thiessen\\
   University of Tübingen and Tübingen AI Center\\
  Max Planck Institute for Informatics\\
  \texttt{maximilian.thiessen@uni-tuebingen.de}\\
}

\begin{document}

\maketitle

\begin{abstract}
Not all mistakes are alike. For example, in medical diagnosis, false positive and false negative mistakes have a different impact and should not be aggregated. By revisiting the classical theory of boosting through this lens, we introduce the notion of \emph{$Z$-learner}, which returns weak hypotheses achieving joint false-positive/false-negative guarantees $(\FP,\FN)$ from some set $Z \subseteq [0,1]^2$. This formulation generalizes the standard notion of $\gamma$-weak learning---recovered by taking $Z=\{(z_1,z_2): z_1+z_2<1/2-\gamma\}$---and allows for more flexible notions of weak learnability that account for the impact of the distribution on the achievable learning guarantees. This raises a natural question: which sets Z are \emph{boostable}? We answer this by giving a complete geometric characterization that unifies, as special cases, both the classical boosting dichotomy and the recent multi-objective extension of Bressan et al.~(COLT 2025). We also present an adaptive algorithm for boosting $Z$-learners without relying on the prior knowledge of~$Z$, generalizing AdaBoost's ability to handle unknown margin parameters.
Finally, we demonstrate that $Z$-learners can achieve confidence amplification. However, the analysis is significantly more involved than in classical boosting and relies on the centerpoint theorem from discrete and computational geometry.

\end{abstract}

\section{Introduction}

In many real-world prediction tasks, errors naturally split into two distinct types: \emph{false positives} (Type~I errors), where a negative instance is incorrectly labeled as positive, and \emph{false negatives} (Type~II errors), where a positive instance is incorrectly labeled as negative. In practice, these two errors are rarely symmetric. For example, in medical diagnostics, one can often reduce false negatives by using more tests or lowering the decision threshold, but this typically leads to increasing false positives. Conversely, stricter decision criteria reduce false positives at the cost of more false negatives.

Despite the ubiquity of this tension, its theoretical understanding remains limited. A basic illustration of this gap appears in the theory of boosting, a central paradigm in machine learning. In its classic form, boosting guarantees that any weak PAC learner---namely, an algorithm achieving total error at most $1/2 - \gamma$ for some $\gamma > 0$---can be converted into a strong learner with arbitrarily small error.
However, this guarantee treats all mistakes uniformly. Once we distinguish between false positives and false negatives, new and qualitatively different notions of weak learners arise. For instance, consider a learner whose total error may even exceed $1/2$, but which guarantees an extremely small false-positive rate (e.g., at most $0.001$). More generally, one can imagine learners that behave asymmetrically across distributions: for some distributions they achieve a very small false-positive rate, while for others they achieve a very small false-negative rate yet their total error exceeds $1/2$. Such learners would be useless under the standard boosting framework, yet they capture meaningful and practically relevant forms of reliability. This raises a natural question: can such asymmetric guarantees be amplified? More broadly, how should boosting be formulated when the goal is to control false positives and false negatives separately, rather than just their sum?

In order to answer these questions we take a more refined view of weak learning, and describe the learner's guarantee not only by its total error, but by the pairs $(\FP,\FN)$ it achieves.
To this end, we simply encode the learner's guarantees by a generic set $Z \subseteq [0,1]^2$; this means that, under every realizable distribution, the learner achieves an $(\FP,\FN)$ pair which is dominated from above by some $\bz\in Z$, where $\bz$ may depend on the distribution.
For example, the guarantees of a $\gamma$-weak PAC learner in the classical sense are given by the set $Z = \{\bz \in [0,1]^2 : z_1+z_2 \le \frac{1}{2}-\gamma\}$; indeed, under every distribution the learner has total error rate $\FP+\FN \le z_1+z_2 \le \frac{1}{2}-\gamma$.
The classic boosting theorem shows that this guarantee is boostable, for instance by AdaBoost~\citep{Freund97decision}. On the other hand, a guarantee of total error $1/2$ is trivial, since it can be achieved by tossing a fair coin, and is therefore not boostable.
As another example, ~\cite{bressan2025dice} recently considered the case where the learner's guarantees is given by the singleton set $Z = \{\bz\}$; this means that the learner's hypotheses always achieve $\FP \le z_1$ and $\FN \le z_2$.
The results of~\cite{bressan2025dice} tell that such a guarantee is boostable if $\sqrt{z_1} + \sqrt{z_2} < 1$, and not boostable if $\sqrt{z_1} + \sqrt{z_2} \geq 1$, as in that case a $Z$-learner can be implemented via biased random coins.
Our goal is to generalize these results to arbitrary $Z \subseteq [0,1]^2$, which leads to the central question of this work:\begin{center}
\emph{Which sets $Z \subseteq [0,1]^2$ of $(\FP,\FN)$-error guarantees can be boosted?}
\end{center}
Previous work falls short of answering this question.
Moreover it is not clear, in principle, that there is still a sharp boundary between boostable and non-boostable guarantees.
Indeed, there may be intermediate regimes, in which the weak learner cannot be boosted all the way to strong learning, but its guarantee can nevertheless be improved in some nontrivial sense, for example by reducing one type of error or by obtaining a better nonzero error pair.

Our main result shows that the answer is governed by a clean dichotomy: no intermediate regime exists. For every set $Z$, either the guarantee is strong enough to be boosted to arbitrarily high overall accuracy, or it is not more informative than random guessing.
This dichotomy generalizes simultaneously both the classic boosting setting, $Z=\{\bz:z_1+z_2 \le 1/2-\gamma\}$, and the singleton case $Z=\{\bz\}$ of~\cite{bressan2025dice}.
Our result also gives a simple geometric interpretation of the dichotomy: the boostability of $Z$ is determined by whether the convex hull of $Z$, denoted $\conv(Z)$, intersects a certain set $\C \subseteq [0,1]^2$, namely
\begin{equation}
\label{eq:coin_region_def}
\C=\{(z_1,z_2) : \sqrt{z_1} + \sqrt{z_2}\geq 1\}.
\end{equation}
Interestingly, the classical edge (or margin) parameter $\gamma$, which governs the complexity of boosting, admits a natural geometric interpretation in our framework: it corresponds to the $\ell_\infty$ distance between $\conv(Z)$ and the set $\mathcal{C}$. In particular, for the classical choice of $Z$, this notion recovers the standard definition of the boosting margin.

The above characterization identifies exactly when boosting is possible, but it also raises a natural algorithmic question: can boosting be carried out \emph{adaptively}, without prior knowledge of the weak-learning guarantee? For example, in the standard case of $\gamma$-PAC learners,
AdaBoost successfully boosts without knowing in advance the set $Z$ (which, in this case, boils down to knowing $\gamma$). In other words, AdaBoost does not need knowledge of the weak learner's guarantees, nor to know whether the weak-learning condition holds; instead, it automatically identifies and amplifies whatever advantage the weak learner provides.

Our second main result shows that this principle also extends beyond standard boosting. We give a universal boosting algorithm for $(\FP,\FN)$ guarantees that requires no prior knowledge of $Z$, of an underlying hypothesis class, or of the geometric margin separating $\conv(Z)$ from $\C$. Instead, the algorithm identifies the relevant parameters of the weak-learning guarantee during its run, and uses this information to boost whenever boosting is possible. Thus, the clean geometric dichotomy above is accompanied by an adaptive algorithmic counterpart: every boostable $(\FP,\FN)$ guarantee can be boosted without knowing in advance the parameters that certify its boostability.

Our third contribution concerns the amplification of the $Z$-learner's confidence parameter; i.e., the probability that the learner outputs a weak hypothesis whose $(\FP,\FN)$ guarantees are consistent with~$Z$. In the classical PAC setting, this is relatively straightforward: one can run the weak learner several times and use a validation step to select a good hypothesis. In our setting, however, this is more delicate. The learner is required to output a hypothesis whose error vector is consistent with a target set \(Z\), but the algorithm does not know \(Z\), and therefore cannot directly validate which calls succeeded. Our approach to amplification assumes the $Z$-learner is correct with probability larger than $2/3$. The idea is to make order of $\ln(1/\delta)$ calls to the weak learner in each round, where $\delta$ is the desired confidence level, so that with probability at least $1-\delta$ more than $2/3$ of the error vectors returned by these calls are valid for $Z$. This in turn allows us use the notion of centerpoint (an efficiently computable high-dimensional notion of median) of these vectors to construct a randomized weak hypothesis whose corresponding error vector is valid for $Z$.

\subsection{Main Results}\label{subsec:results}
\begin{figure}
    \centering
\begin{tikzpicture}[scale=3.7]
    \small
    \def\maxx{.6};
    \def\maxy{.6};
    \draw[thick,-latex] (0,0) -- (0,\maxy+.05);
    \draw[thick,-latex] (0,0) -- (\maxx+.05,0);
    \node at (-.06,\maxy) {$z_2$};
    \node at (\maxx,-.05) {$z_1$};
    \clip (-.2,-.2) rectangle (\maxx,\maxy);

    \draw[thick,domain=0:1,samples=50] plot (\x,{(1-sqrt(\x))^2});
    \node[] at (.4,.4) {$\C$};
    \fill[opacity=.15,domain=0:1,samples=50] plot (\x,{(1-sqrt(\x))^2});
    \fill[opacity=.15] (0,1) -- (1,0) -- (1,1) -- cycle;

    \coordinate (a) at (0,.47);
    \coordinate (b) at (.47,0);
    \node[left=0pt] at (a) {\scriptsize $\frac{1}{2}$-$\gamma$};
    \node[below=0pt] at (b) {\scriptsize $\frac{1}{2}$-$\gamma$};
    \draw[-,name path = A,thick, cyan] (a) -- (b);
    \fill[cyan,opacity=.1] (0,0) -- (a) -- (b) -- (1,0) -- cycle;

\end{tikzpicture}
\begin{tikzpicture}[scale=3.7]
    \small

    \def\maxx{.6};
    \def\maxy{.6};
    \draw[thick,-latex] (0,0) -- (0,\maxy+.05);
    \draw[thick,-latex] (0,0) -- (\maxx+.05,0);
    \node at (-.06,\maxy) {$z_2$};
    \node at (\maxx,-.05) {$z_1$};
    \clip (-.2,-.2) rectangle (\maxx,\maxy);

    \draw[thick,domain=0:1,samples=50] plot (\x,{(1-sqrt(\x))^2});
    \node[] at (.4,.4) {$\C$};
    \fill[opacity=.15,domain=0:1,samples=50] plot (\x,{(1-sqrt(\x))^2});
    \fill[opacity=.15] (0,1) -- (1,0) -- (1,1) -- cycle;

    \coordinate (a) at (.17,.3);
    \draw[-,thick, cyan] (0,0 -| a) -- (a) -- (a -| 0,0);
    \fill[cyan,opacity=.1] (0,0 -| a) -- (a) -- (a -| 0,0) -- (0,0) -- cycle;
    \fill[cyan] (a) circle (.015);

\end{tikzpicture}
\begin{tikzpicture}[scale=3.7]
    \small

    \def\maxx{.6};
    \def\maxy{.6};
    \draw[thick,-latex] (0,0) -- (0,\maxy+.05);
    \draw[thick,-latex] (0,0) -- (\maxx+.05,0);
    \node at (-.06,\maxy) {$z_2$};
    \node at (\maxx,-.05) {$z_1$};
    \clip (-.2,-.2) rectangle (\maxx,\maxy);

    \draw[thick,domain=0:1,samples=50] plot (\x,{(1-sqrt(\x))^2});
    \node[] at (.4,.4) {$\C$};
    \fill[opacity=.15,domain=0:1,samples=50] plot (\x,{(1-sqrt(\x))^2});
    \fill[opacity=.15] (0,1) -- (1,0) -- (1,1) -- cycle;

    \def\dotsize{0.015}
    \coordinate (a1) at (.1,.27);
    \coordinate (a2) at (.3,.17);
    \coordinate (a3) at (.45,.05);
    \coordinate (a4) at (.15,.18);
    \coordinate (a5) at (.25,.07);
    \draw[-,thick, cyan] (0,0 |- a1) -- (a1) -- (a2) -- (a3) -- (a3 |- 0,0);
    \fill[cyan,opacity=.1] (0,0 |- a1) -- (a1) -- (a2) -- (a3) -- (a3 |- 0,0) -- (0,0) -- cycle;

    \foreach \i in {1,...,5} {
        \fill[cyan] (a\i) circle (\dotsize);
    }

\end{tikzpicture}
\begin{tikzpicture}[scale=3.7]
	\small

	\def\maxx{.6};
	\def\maxy{.6};
	\draw[thick,-latex] (0,0) -- (0,\maxy+.05);
	\draw[thick,-latex] (0,0) -- (\maxx+.05,0);
	\node at (-.06,\maxy) {$z_2$};
	\node at (\maxx,-.05) {$z_1$};
	\clip (-.2,-.2) rectangle (\maxx,\maxy);

	\draw[thick,domain=0:1,samples=50] plot (\x,{(1-sqrt(\x))^2});
	\node[] at (.4,.4) {$\C$};
	\fill[opacity=.15,domain=0:1,samples=50] plot (\x,{(1-sqrt(\x))^2});
	\fill[opacity=.15] (0,1) -- (1,0) -- (1,1) -- cycle;

	\def\dotsize{0.015}

		\coordinate (a1) at (0, .5);
		\coordinate (a2) at (.5, 0);
        \coordinate (overlap) at (.25, .25);

	\draw[-,thick,cyan,dotted] (0,0 |- a1) -- (a1) -- (a2) -- (a2 |- 0,0);

    \foreach \i in {1,...,2} {
        \fill[cyan] (a\i) circle (\dotsize);
    }
    \node[star,fill=red,draw=red,star points=4,star point height=1.7,minimum size=2,inner sep=0.6pt,star] at (overlap) {};
\end{tikzpicture}
     \caption{A visual depiction of our dichotomy theorem. In each plot, the two axes show the false-positive error guarantee $z_1$ and the false negative error guarantee $z_2$. The gray region is the set of ``trivial'' (coin-attainable) guarantees $\C$ (see \cref{eq:coin_region_def}). The blue region is (the down-closure of) some set of $(\FP,\FN)$ error guarantees $Z \subseteq [0,1]^2$ . In the left plot $Z=\{\bz\in[0,1]^2:z_1+z_2 \le 0.5-\gamma\}$, which is boostable by classic results. In the second plot $Z=\{\bz\}$ with $\bz \in [0,1]^2 \setminus \C$, which is boostable by the results of~\cite{bressan2025dice}. In the third plot $Z$ is any subset of $[0,1]^2$ with $\conv(\overline{Z})\cap \C = \emptyset$; \Cref{thm:main} says this condition characterizes  the boostable sets, therefore subsuming the two previous cases. In the rightmost plot, $Z=\{(0,\nicefrac{1}{2}),(\nicefrac{1}{2},0)\}$; while the two marked points are individually boostable, the set $Z$ is not as its convex hull touches $\C$.}
    \label{fig:plots}
\end{figure}
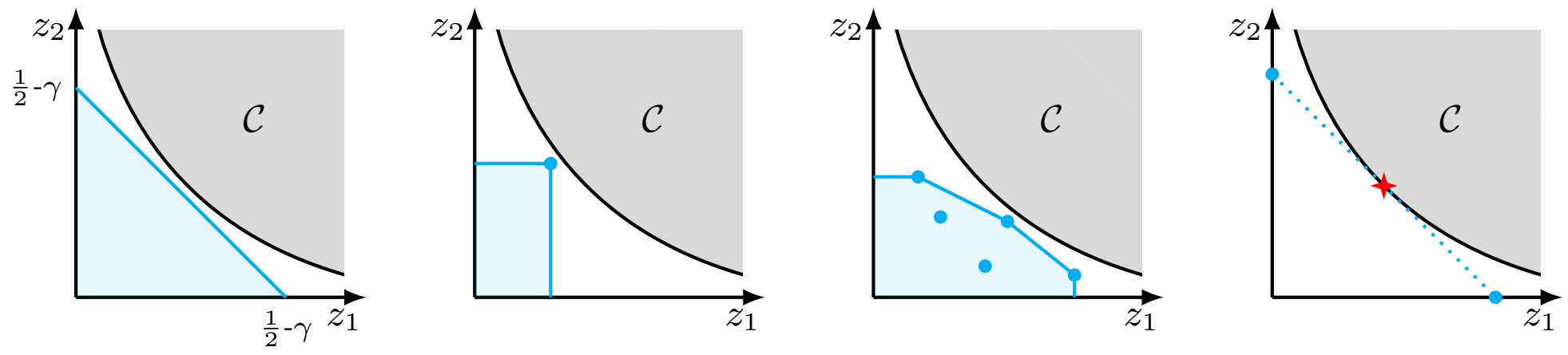

Fix a domain $\X$, let $\Y=\{-1,+1\}$, and let $\F \subseteq \Y^\X$ be a concept class. For a target function $f \in \F$, a distribution $\D$ over $\X$, and a hypothesis $h: \X \to \Y$, we distinguish between two types of errors: false positives and false negatives. The false-positive (FP) error measures how often $h$  predicts $+1$ on negative examples, while the false-negative (FN) error measures how often $h$ predicts $-1$ on positive examples. Formally,
\begin{equation}\label{eq:fp}
 e_1 := \Pr_{x \sim \D}\left[h(x) = 1 \land f(x) = -1\right], \qquad
 e_2 := \Pr_{x \sim \D}\left[h(x) = -1 \land f(x) = 1\right].
\end{equation}
We denote the vector $(e_1,e_2)$ by $\errs_{\D,f}(h)$, and refer to it as the \emph{error vector} of $h$.\footnote{We will overload notation when $\D$ is a distribution over $\X \times \Y$, writing $\errs_{\D}(h)$ for the corresponding errors.}
When $h$ is a randomized hypothesis, and thus the prediction $h(x)$ is a random variable, the probabilities above are meant as taken over $h$ as well.

In order to state our boostability results, we need to introduce some essential notions.
The first one, which is novel, is the notion of $Z$-learner. As said above, the idea is to measure the performance not by the total error, but by the set of $(\FP,\FN)$ pairs achieved.
For $\bz=(z_1,z_2)\in\RR^2$ and $\epsilon\in\RR$ we will denote
$\bz+ \beps = (z_1+\epsilon,z_2+\epsilon)$; inequalities in the form $\bz \le \bz'$ are meant as coordinate-wise.

\begin{definition}[$Z$-learner]\label{def:Z-learner}
Let $Z \subseteq [0,1]^2$.
An algorithm~$\A$ is a $Z$-learner for a concept class $\F$ if there exists a function $m_0: (0,1)^2 \rightarrow \mathbb{N}$ such that for every $f \in \F$, every distribution $\D$ over $\X$, and every $\epsilon, \delta \in (0,1)$ the following holds.
If $S$ is a sample of $m_0(\epsilon, \delta)$ examples drawn i.i.d.\ from $\D$ and labeled by $f$, then $\A(S)$ returns a predictor $h: \X \rightarrow \Y$ such that, with probability at least $1-\delta$,
\[
\exists \bz \in Z \;:\; \errs_{\D,f}(h) \le \bz +  \beps\,.
\]
\end{definition}
As discussed above, $Z$-learners subsume and generalize both the standard notion of $\gamma$-weak PAC learner and of $\bz$-learner.
Our goal is to understand for which $Z$ every $Z$-learner can be boosted to a strong learner (i.e., a PAC learner).
To this end we need to define what ``boostable'' means.
For two algorithms \(\A\) and \(B\), we denote by \(B^\A\) the algorithm \(B\) equipped with black-box oracle access to \(\A\).
\begin{definition}[Boostable $Z$]
A set \(Z \subseteq [0,1]^2\) is \emph{boostable} if there exists an algorithm \(B\), called \emph{boosting algorithm}, with the following property: for every concept class \(\F \subseteq \Y^\X\) and every $Z$-learner \(\A\) for~\(\F\), the algorithm \(B^\A\) is a PAC learner for \(\F\).
\end{definition}
At the other end of the spectrum, we deem a guarantee $Z$ trivial if it can be achieved by randomized predictors that ignore the input entirely, as in~\cite{bressan2025dice}.
Denote by $\Delta_\Y$ the family of all distributions over $\Y$ (i.e., the $1$-dimensional standard simplex).
\begin{definition}[Trivial $Z$]\label{def:random-guess}
A (randomized) hypothesis $h$ is a \emph{random guess} if there exists a distribution $\bp \in \Delta_\Y$ such that $h(x) \sim \bp$ independently for every $x \in \X$.\footnote{It is not needed for our results that every \emph{evaluation} of $h$, even on the same $x$, follows $\bp$ independently.}
A $Z \subseteq [0,1]^2$ is \emph{trivial} if there exists an algorithm that outputs only random guesses and is a $Z$-learner for every concept class $\F \subseteq \Y^\X$.
\end{definition}
Obviously, if $Z$ is trivial then $Z$ is not boostable, since there exists a $Z$-learner that outputs only random guesses, and no aggregation of random guesses can yield a strong learner.
Conversely, if $Z$ is boostable then it cannot be trivial.

We are now ready to state our first main result: an explicit characterization of each set $Z$ as either boostable or trivial.
Recall the set $\C = \{(z_1,z_2) \in [0,1]^2 : \sqrt{z_1} + \sqrt{z_2} \ge 1\}$.
For any $Z \subseteq \RR^2$, we denote by $\overline{Z}$ its topological closure and by $\conv(Z)$ its convex hull.
\begin{theorem}[{\bf Boostability dichotomy for $Z$-learners}]\label{thm:main}
Let $Z \subseteq [0,1]^2$. Then:
\begin{enumerate}[itemsep=-1ex,topsep=0pt]
    \item If ${\conv\big(\overline{Z}\big)} \cap \C = \emptyset$, then $Z$ is boostable.
    \item Otherwise (i.e., if $\conv\big(\overline{Z}\big) \cap \C \ne \emptyset$), $Z$ is trivial.
\end{enumerate}
\end{theorem}
For the sake of simplicity, in what follows we assume the set $Z$ is closed, i.e., $\overline{Z} = Z$. However, all of our results do apply to the general case as well.

As noted above, \Cref{thm:main} yields as special cases both the classic boostability dichotomy for standard weak PAC learners, corresponding to guarantees in the form $Z=\{\bz \in [0,1]^2:z_1+z_2 \le 1/2-\gamma\}$, and the recent boostability dichotomy of~\cite{bressan2025dice} for the special case of singleton guarantees, $Z=\{\bz\}$.
We remark that \Cref{thm:main} is a genuine generalization; it cannot be obtained as a direct consequence or reinterpretation of previous results (or their proofs), and those results are truly special cases of it.

The last two plots of \Cref{fig:plots} give two concrete examples that are captured by \Cref{thm:main} but are not captured by previous results.
The third plot shows a set $Z\subseteq[0,1]^2$ that satisfies $\conv(Z) \cap \C = \emptyset$, and is thus boostable by \Cref{thm:main}; however, this set is not deemed boostable by previous results    (for instance, it does not identify a standard weak PAC learner).
The fourth plot shows the set $Z=\{(0,\nicefrac{1}{2}),(\nicefrac{1}{2},0)\}$. Individually, both $\{(0,\nicefrac{1}{2})\}$ and $\{(\nicefrac{1}{2},0)\}$ are boostable by the results of \citet{bressan2025dice}.
At first sight, one might expect that this implies that their union $Z$ is boostable, too.
Instead, \Cref{thm:main} says that $Z$ is not boostable, as $\conv(Z)$ intersects $\C$ in  $(\nicefrac{1}{4},\nicefrac{1}{4})$; and this case, too, is not captured by previous results.

Our second main result is to show that, in fact, $Z$-learners can be boosted \emph{adaptively}.
Adaptivity is a key feature of classic algorithms such as AdaBoost: it means those algorithms do not require prior knowledge of the weak learner's guarantees, but rather exploit automatically whatever advantage is present.
We show that this phenomenon extends to the case of $Z$-learners by giving a \emph{universal} boosting algorithm---one that requires no prior knowledge of $Z$, of the concept class $\F$, or of the geometric margin separating $\conv(Z)$ from $\C$.
\begin{theorem}[{\bf Adaptive Boosting}]\label{thm:adaptive}
There exists a boosting algorithm $B$ such that for any boostable $Z \subseteq [0,1]^2$, class $\F \subseteq \Y^\X$, and $Z$-learner $\A$ for $\F$, the algorithm $B^\A$ is a PAC learner for $\F$.
\end{theorem}
The high-level idea behind the algorithm of \Cref{thm:adaptive} is to find a  {line} that separates $\conv(\bar{Z})$ from $\C$.
The normal vector $\bw \in \Delta_\Y$ corresponding to this  {line} can then be used to boost, as shown in~\cite{bressan2025dice}.
The difficulty is that we do not know $Z$ in advance, and we have only black-box access to the $Z$-learner $\A$.
We show that we can nonetheless find $\bw$ with a small overhead, by starting with an initial guess $\bw^{(1)}$, and testing along the boosting procedure for the success of $\A$ in providing a hypothesis that is better than a random guess.
When $\A$ fails, we show that we can gather enough information to perform a binary search towards $\bw$.
The resulting algorithm is also efficient: the number of calls to the weak learning algorithm $\A$ as well as its sample complexity are within small factors of the ones of AdaBoost.

One aspect of the algorithm in \Cref{thm:adaptive}, which is also interesting in its own right, is that it requires the weak learner to succeed with high confidence: its failure probability $\delta$ must be small as a function of the number of oracle calls made during boosting. This raises the question of whether such high confidence can be obtained from a weak learner that only has fixed confidence, again without knowing~$Z$. Our next result answers this question affirmatively.
To formalize this, we introduce the following notion. Let $\delta_0>0$. We say that $\A$ is a $(\delta_0,Z)$-learner for $\F$ if it satisfies the definition of a $Z$-learner (\Cref{def:Z-learner}), except that the confidence parameter is fixed to be $\delta = \delta_0$.
\begin{theorem}[{\bf Confidence Amplification}]\label{thm:confidence-amplification-intro}
There exists an algorithm $B$ such that for every $Z \subseteq [0,1]^2$, every concept class $\F \subseteq \Y^\X$, and every $(\delta_0,Z)$-learner $\A$ for $\F$ with $\delta_0<1/3$, the algorithm $B^\A$ is a $\conv(Z)$-learner for $\F$.
\end{theorem}
Moreover, the amplification incurs only a multiplicative $O(\log(1/\delta))$ overhead in the relevant complexity measures, where $\delta$ is the desired failure probability; see \Cref{sec:confboost} for more details.

\paragraph{Technical challenges.}
Proving our results requires several new ingredients missing from both classical adaptive boosting results \citep{schapire2012boosting} and from the recent generalized boosting setting of \citet{bressan2025dice}.
In particular, the weak learner's guarantee is not a fixed point $\bz$, but rather a possibly distribution-dependent error vector lying in the downward closure of an \emph{unknown} set $Z$.
Consequently, the margin depends on the separation between $\conv(Z)$ and $\C$.
This makes adaptivity substantially harder than in AdaBoost: in addition to building an appropriate ensemble of predictors (as in standard boosting approaches), the algorithm must recover an unknown separating direction $\bw^*$ and the respective margin $\gamma^*$ from black-box calls to the weak learner.
The confidence amplification argument is also much more delicate given that $Z$ is unknown, requiring further technical results from discrete geometry that are not needed in the simpler boosting settings from prior work.
Finally, the lower bound requires a novel construction based on distribution-dependent random guesses, showing that even adaptive random coins (required by the nature of our more general $Z$-learner guarantee) can satisfy the $Z$-learner condition whenever $\conv(Z) \cap \C \neq \emptyset$, which is a new key step in proving the necessity of the separation between $\conv(Z)$ and $\C$.

\subsection{Related Work}
This work follows a long thread of research on the boostability of weak learners and the corresponding adaptive algorithms. For an overview on boosting theory see the book by \citet{schapire2012boosting}. Our main subroutine (\Cref{alg:binary_boost}), which we adapted from \citet{bressan2025dice}, can be seen as a variant of AdaBoost \citep{Freund97decision}. Many similar variants of AdaBoost exist \citep{nikolaou2016cost, landesa2012shedding,viola2001fast} that explicitly take different false positive and false negatives errors into account; often studied from the adjacent perspective of cost-sensitive learning. In fact, the weight update of \Cref{alg:binary_boost} matches exactly the modification of AdaBoost by \citet{sun2007cost}. However, such previous results fall short of characterizing the boostability of weak learners with guarantees in a set $Z\subseteq [0,1]^2$, as we do.

More broadly, cost-sensitive classification has a long
history of studying prediction under asymmetric
misclassification costs. Classical approaches include
cost-sensitive decision rules~\cite{elkan2001foundations},
meta-learning methods~\cite{domingos1999metacost},
and reductions based on example
reweighting~\cite{zadrozny2003cost}.
Further theoretical work studies the calibration
of asymmetric surrogate losses~\cite{scott2012calibrated},
while Masnadi-Shirazi and Vasconcelos~\cite{masnadi2011cost}
developed principled cost-sensitive boosting algorithms.
Unlike these approaches, which typically optimize a
specified cost-sensitive objective, our work studies
the boostability of weak learners whose achievable
(FP, FN) guarantees may vary across distributions.
Our characterization determines when such guarantees
can be amplified to arbitrarily small error, even
without knowing the underlying trade-off.

Our results can also be placed into the broader context of learning with multi-objective losses \citep{jin2006multi,cortes2020agnostic}. For example, our weak learning guarantees with a set $Z\subseteq [0,1]^2$ encompass \emph{reliable} learning \citep{kalai2012reliable} and learning with \emph{one-sided error} \citep{kivinen1995learning}.
\section{Preliminaries}
Let $\X$ be any domain and $\Y=\{-1,+1\}$.
An algorithm is a PAC learner for $\F \subseteq \Y^\X$ if there is a function $m: (0,1)^2 \rightarrow \mathbb{N}$ such that for every $f \in \F$, every distribution $\D$ over $\X$, and every $\epsilon,\delta \in (0,1)$ the following claim holds. If $S$ is a sample of $m(\epsilon, \delta)$  examples drawn i.i.d.\ from $\D$ and labeled by $f$, then $\A(S)$ returns a predictor $h: \X \rightarrow \Y$ such that, with probability at least $1-\delta$, $\Pr_{x \sim \D}[{ h(x) \neq  f(x)}] \le \epsilon$.

A \emph{weight} is any function $\bw:\Y^2 \to [0,1]$ that satisfies $\bw(i,i)=0$ for all $i \in \Y$.  For conciseness, we let $w_1 = \bw(+1,-1)$ and $w_2 = \bw(-1,+1)$, and by some overloading of notation we denote by $\bw$ both the matrix $\bigl(\begin{smallmatrix} 0  & w_2\\ w_1 & 0 \end{smallmatrix}\bigr)$ and the vector $(w_1,w_2)$. Throughout the paper we will generally require that $w_1 + w_2 = 1$.

A point $\bz = (z_1,z_2) \in [0,1]^2$ is \emph{coin-attainable} if for every $\bq \in \Delta_\Y$ there exists $\bp \in \Delta_\Y$ such that $p_+ q_- \le z_1$ and $p_- q_+ \le z_2$.
This means that the random guess $h_{\bp}$ satisfies $\errs_{\D,f}(h_{\bp}) \le \bz$ for every distribution $\D$ with label marginal $\bq$ and every function $f:\X\to\Y$.
The set of all coin-attainable points is denoted by $\C$. Formally,
\begin{align}\label{eq:C_as_coinatt}
    \C = \left\{ (z_1,z_2) \in [0,1]^2 : \forall \bq \in \Delta_\Y  \ \exists \bp \in \Delta_\Y \ \text{ s.t. } \ p_+ q_- \le z_1 \ \land \  p_- q_+ \le z_2\right\}.
\end{align}
As shown in \cite{bressan2025dice}, the set $\C$ can be equivalently defined as \cref{eq:coin_region_def}.

The \emph{margin} of a guarantee set $Z \subseteq [0,1]^2$ measures its separation from the coin-attainable set $\C$.
Formally, it is the $\ell_\infty$ distance between $\C$ and $\conv(Z)$:
\begin{equation}
    \gamma^* :=
\min_{\substack{\bz'\in \conv(Z)\\ \bz\in \C}}
\|\bz-\bz'\|_\infty.
\end{equation}
Provided that ${\conv\big(\overline{Z}\big)} \cap \C = \emptyset$, this is equivalent to measuring distance along a maximizing normal direction, that is,
\begin{equation}\label{eq:margin_def}
    \gamma^* := \max_{\bw\in\Delta_2}\min_{\bz\in\conv(Z)}\left(w_1w_2 - \bw\cdot \bz\right).
\end{equation}
The proof of the equivalence relies on standard norm duality and is deferred to \Cref{subsec:proof:margin_equivalence} and, more precisely, to \Cref{lem:margin_equivalence}.
Thus,  by \cref{eq:margin_def}, $\gamma^*$ measures the maximum gap between the projections of $\C$ and $\conv(Z)$ along a normal direction $\bw \in \Delta_2$.
We denote by $\bw^* \in \Delta_2$ the direction that attains this maximum (formally, $\bw^*$ is the argument of the $\max$ in \cref{eq:margin_def}).

One goal of our adaptive boosting algorithm is to identify a good estimate of $\bw^*$.
Since $\bw^* \in \Delta_2$, it suffices to consider its first element $w_1^*$ alone (as $w_2^* = 1- w_1^*$).
For every $\bz = (z_1,z_2) \in \conv(Z)$ define the function $v(\cdot\,;\bz) : \RR \to \RR$ by letting:
\begin{equation}\label{eq:def_v_valid}
     v(w_1; \bz) :=w_1(1-w_1)-(w_1z_1+(1-w_1)z_2).
\end{equation}
It is immediate to see that the right-hand expression of \Cref{eq:def_v_valid} is just the inner expression of \Cref{eq:margin_def}, and therefore:
\begin{equation}
\gamma^*=\max_{w_1\in[0,1]} \min_{\bz\in\conv(Z)} v(w_1; \bz)
\end{equation}
In particular, $v(w_1^*; \bz) \ge \gamma^*$ for every $\bz\in \conv(Z)$.
We will refer to $v(\cdot\,; \cdot)$ as the \emph{validity} function, as it verifies the validity of $w_1$ with respect to a given point $\bz$.
The validity function is used by our boosting algorithm in \Cref{sec:boost} to find good estimates $w_1,\gamma$ of $\w_1^*,\gamma^*$.

\section{Adaptive Boosting}\label{sec:boost}
In this section we give our main  adaptive boosting algorithm, whose formal guarantees are stated in
\Cref{thm:adaptive_boost_formal}.
The algorithm pseudocode is given in \Cref{alg:adaptive} and discussed below.
The algorithm needs a good estimate $\gamma$ of the true margin $\gamma^*$ of the $Z$-learner;
this assumption can be removed by a standard guessing-by-halving strategy, as discussed later on.
Similarly, the algorithm requires quite a strong success probability of the weak learner; this requirement can be relaxed, too, by the confidence boosting result of \Cref{sec:confboost}.

\begin{theorem}[Adaptive Boosting] \label{thm:adaptive_boost_formal}
Let $\F \subseteq \Y^\X$, and let $\A$ be a $Z$-learner for $\F$ with margin $\gamma^* > 0$ and sample complexity $m_0$. Fix any $\epsilon, \delta \in (0,1)$.
Let $f \in \F$, let $\D$ be any distribution over $\X$, and let $S=\{(x_1,y_1),\ldots,(x_m,y_m)\}$ be a sample with $x_1,\dots,x_m \sim \D$ drawn i.i.d.\ and labeled by $f$.
If \Cref{alg:adaptive} is given oracle access to $\A$, parameters  $\gamma \le \gamma^*/2$, $R=\left\lceil \log(\tfrac{3}{\gamma})\right\rceil$, $T = \ceil*{\frac{18\ln(m)}{\gamma^2}}$, $\eta = \sqrt{\frac{2\ln(m)}{T}}$, and $\widehat m = m_0\bigl(\frac{\gamma}{3}, \tfrac{\delta}{2TR} \bigr)$, and a sample $S$ of size $m \ge \Tilde{O}\!\left(\tfrac{\widehat m}{\epsilon\gamma^2}\right)$, then \Cref{alg:adaptive} makes
$TR$ calls to $\A$ and returns a (possibly randomized) predictor $\widehat h_S:\X\to\Y$ such that with probability at least $1-\delta$,
\begin{equation*}
    \Pr_{x \sim \D}\Big[\widehat h_S(x) \neq f(x) \Big] \le \epsilon.
\end{equation*}
\end{theorem}

The idea behind \Cref{alg:adaptive} is to proceed in rounds, as follows. At any round $r=1,2,\ldots$, the algorithm maintains an interval \([L_r,U_r] \subseteq [0,1]\) guaranteed to contain the first component \(w_1^*\) of the optimal vector $\bw^*$.
The algorithm then picks the centerpoint of the interval, call it $w_1^{(r)}$, and tries to boost $\A$ over the sample $S$ using the cost-sensitive binary boosting algorithm of~\cite{bressan2025dice} using $\bw^{(r)}=(w_1^{(r)},\,1-w_1^{(r)})$ as cost function.
If boosting succeeds, meaning that the algorithm of~\cite{bressan2025dice} yields a hypothesis $\hat{h}_S$ consistent with the sample $S$, then we are done; $\hat{h}_S$ then achieves zero error on $S$, and by compression-based generalization bounds achieves error at most $\epsilon$ on the original distribution $\D$ as well.
If instead boosting fails, we prove that the weak hypotheses returned by the $Z$-learner, together with our validity function $v$ (\cref{eq:def_v_valid}), can be used to decide whether $w_1^* < w_1^{(r)}$ or $w_1^* > w_1^{(r)}$.
In this way we can halve the interval $[L_r,U_r]$, thus effectively yielding a binary search for the value $w_1^*$.
More precisely, we can prove the algorithm stops when the size of the interval $[L_r,U_r]$ drops below $\gamma$, which happens after $O(\log(1/\gamma))$ iterations.

\begin{algorithm}
\caption{Adaptive generalized boosting} \label{alg:adaptive}
\begin{algorithmic}[1]\onehalfspacing
\REQUIRE sample $S = (x_i,y_i)_{i=1}^m$; weak learner $\A$; parameters $R, T, \eta, {\widehat m}, \gamma$.
\STATE Initialize $L_1=0, U_1 = 1$.
\FOR{$r = 1, \ldots, R$}
\STATE Set \(w_1^{(r)} = \tfrac12(L_r+U_r)\), and set $\bw^{(r)}= (w_1^{(r)},1- w_1^{(r)})$.
\STATE  Call \Cref{alg:binary_boost} with weights $\bw^{(r)}$, and parameters above.
    \IF{\Cref{alg:binary_boost} returned $\widehat h_S$}
        \STATE \textbf{return} $\widehat h_S$.\label{line:adaptive_return_h_S}
    \ELSE
        \STATE \Cref{alg:binary_boost} returned \(\bz\) such that \(v(w_1^{(r)}; \bz)<\gamma\). \hfill \textcolor{gray!80}{ (see \cref{eq:def_v_valid})}
        \label{step:witness}
        \STATE Set \(\Delta_r = 1-2w_1^{(r)}-z_1 + z_2\).
        \STATE \textbf{if} {\(\Delta_r\ge 0\)} \textbf{then} update \(L_{r+1}= w_1^{(r)}\), \(U_{r+1}= U_r\).
        \STATE \textbf{if} {\(\Delta_r<0\)} \textbf{then} update \(L_{r+1}= L_r\), \(U_{r+1}= w_1^{(r)}\).
    \ENDIF
\ENDFOR
\end{algorithmic}
\end{algorithm}

\begin{algorithm}[ht]
\caption{Boosting a weak learner given $\bw$}
\label{alg:binary_boost}
\begin{algorithmic}[1]\onehalfspacing
\REQUIRE sample $S = (x_i,y_i)_{i=1}^m$; weak learner $\A$; weights $\bw = (w_1, w_2)$, parameters $n, T, \eta, \widehat{m}, \gamma$.
\STATE Initialize: $D_1(i) = 1$  for all $i=1,\dots,m$.
\FOR{$t = 1, \ldots, T$}
\STATE Compute the distribution $\D_t \triangleq \frac{D_t}{\sum_{i}D_t(i)}$ over $[m]$.
\STATE Draw a set $S_t$ of $\widehat{m}$ labeled examples i.i.d.\ from $D_t$ and obtain $h_t = \A(S_t)$.
\STATE Let $\bz^{(t)}=\errs_{\D_t,f}(h_t)$.
\IF{$v(w_1; \bz^{(t)}) < \gamma$}
            \RETURN \textbf{fail} with $\bz^{(t)}$.\label{step:fail}
\ENDIF
\STATE  For every $i=1,\ldots,m$,  let $D_{t+1}(i) \triangleq D_t(i) \cdot e^{\eta \cdot\bw(h_t(x_i),y_i)
} \;.$
\ENDFOR
\STATE For all $x \in \X, y\in \Y$, let $F(x,y) \triangleq \frac{1}{T}\sum_{t=1}^T \ind{h_t(x) = y}$.
\RETURN $\widehat h_S: \X \to \Y$ such that, for all $x \in \X$,
\[
\widehat h_S(x) \triangleq
\begin{cases}
+1  & \text{if } w_2 \cdot F(x,-1) < w_1\cdot F(x,1) \;, \\
-1  & \text{otherwise.}
\end{cases}
\]
\end{algorithmic}
\end{algorithm}

The subroutine that attempts the cost-sensitive boosting is given in \Cref{alg:binary_boost}.
The probability of success of \Cref{alg:binary_boost} depends on the input parameters; to simplify the analysis, we first show that the routine is successful, and outputs a hypothesis $\hat{h}_S$ consistent with the sample $S$, if the failure step (line \ref{step:fail}) is never reached.

\begin{restatable}{lemma}{ConsistencyLemma}\label{lem:binary_scalar_boosting}
Let $\F \subseteq \Y^\X$, let $\A$ be a $Z$-learner for $\F$, and let $S=\{(x_1,y_1),\ldots,(x_m,y_m)\}$ be a sample of points labeled by some $f \in \F$.
Suppose that, upon calling \Cref{alg:binary_boost} with $S$, oracle access to $\A$, and parameters $T$, $\eta$, $\widehat m$ as in \Cref{thm:adaptive_boost_formal}, line \ref{step:fail} is never reached.
Then the algorithm makes $T$ calls to $\A$ and returns a hypothesis\footnote{For simplicity, we present \Cref{alg:binary_boost} for weak hypotheses that are deterministic at prediction time (although the weak learner itself may be randomized). The algorithm and analysis extend to randomized weak hypotheses by replacing the losses in the weight updates by their expectations and the indicators in Line 11 by prediction probabilities.} $\widehat h_S:\X\to\Y$ consistent with $S$, that is, such that
\[
     \widehat h_S(x_i) = y_i \qquad \forall i =1,\ldots,m \;.
\]
\end{restatable}

The proof of \Cref{lem:binary_scalar_boosting} follows from Lemma~12 in  \cite{bressan2025dice}, under the condition that step~\ref{step:fail} is never reached.
We can now show that our adaptive algorithm (\Cref{alg:adaptive}), when called with ``sufficiently good'' parameters, will then also find a value $w_1$ that is sufficiently close to $w_1^*$.
Consequently, by \Cref{lem:binary_scalar_boosting}, it returns a hypothesis $\hat{h}_S$ that is consistent with the sample.

\begin{lemma}\label{lemma:zero_trainig_err}
    Let $\F \subseteq \Y^\X$, and let $\A$ be a $Z$-learner for $\F$ with margin $\gamma^*>0$ and sample complexity $m_0$.
Let $f \in \F$, let $\D$ be any distribution over $\X$, and let $S=\{(x_1,y_1),\ldots,(x_m,y_m)\}$ be a sample with $x_1,\dots,x_m \sim \D$ drawn i.i.d.\ and labeled by $f$.
Fix any $\delta \in (0,1)$. If \Cref{alg:adaptive} is given $S$, oracle access to $\A$, and parameters   as in \Cref{thm:adaptive_boost_formal}, then \Cref{alg:adaptive} makes $TR$ calls to $\A$ and returns a  predictor $\widehat h_S:\X\to\Y$ such that with probability at least $1-\delta/2$,
\[
     \widehat h_S(x_i) = y_i \qquad \forall i =1,\ldots,m \;.
\]
\end{lemma}

The proof of \Cref{thm:adaptive_boost_formal} then follows by first applying \Cref{lemma:zero_trainig_err} to obtain a hypothesis $\widehat h_S$ that is consistent with the training data, and then using compression-based generalization arguments to obtain the final claim. The complete  proof of \Cref{thm:adaptive_boost_formal}
is deferred to the appendix.

\paragraph{A remark on adaptivity to $\gamma^*$.}
Note that \Cref{thm:adaptive_boost_formal} requires one to give \Cref{alg:adaptive} a value $\gamma \le \gamma^*/2$.
If one does not know $\gamma^*$, one can use a standard halving trick: invoke \Cref{alg:adaptive} with $\gamma=\nicefrac{1}{2},\nicefrac{1}{4},\nicefrac{1}{8},\ldots$ until success.
This adds only a multiplicative $O(\ln(1/\gamma^*))$ overhead to the running time and the number of oracle calls to the $Z$-learner $\A$, without affecting the sample complexity.\footnote{The additional meta-rounds do not multiply the sample complexity, since they reuse the same labeled sample $S$
 and resample from distributions over its indices. If the required size of $S$
 is not known in advance, one may maintain nested samples and draw additional examples only when the current guess requires them. By round $J$, the total number of distinct examples is therefore the requirement corresponding to $\gamma_J = \Theta(\gamma^*)$, rather than the sum over the preceding rounds. The $O(\ln(1/\gamma^*))$
 overhead applies only to runtime and oracle calls. }

\paragraph{Comparison with classical boosting.}
Classical boosting algorithms such as Adaboost require
$O(\log(m)/\gamma^2)$
weak-learner calls. Our algorithm matches this bound
when the separating direction is known, but incurs
an additional $O(\log(1/\gamma))$ factor to identify
the unknown direction, yielding a total of
$TR=O(\log(m)\log(1/\gamma)/\gamma^2)$ calls.
To ensure that all calls succeed with high probability,
each invocation requires confidence
$1-\delta/(2TR)$, resulting in a sample complexity of
$\widetilde O(m_0(\gamma/3,\delta/(2TR))/(\epsilon\gamma^2))$,
as stated in \Cref{thm:adaptive_boost_formal}.
If the weak learner only provides constant confidence,
our confidence-amplification procedure (discussed in the next section) adds
$O_{\delta_0}(\log(TR/\delta))$ repetitions per call,
together with an independent validation sample.

\section{Confidence Boosting}
\label{sec:confboost}

In this section we prove~\Cref{thm:confidence-amplification-intro} by giving a confidence-boosting procedure that takes as input a \((\delta_0,Z)\)-learner that succeeds with constant probability \(1-\delta_0\), where \(\delta_0<1/3\), and boosts its confidence to arbitrarily small failure probability, while preserving accuracy guarantees.

For simplicity and without loss of generality, we assume that \(Z \subseteq [0,1]^2\) is convex. Indeed, every \(Z\)-learner is in particular a \(\conv(Z)\)-learner, and the conclusion of \Cref{thm:confidence-amplification-intro} is a high-confidence \(\conv(Z)\)-learner. Moreover, by \Cref{thm:main}, replacing \(Z\) with \(\conv(Z)\) does not affect whether it is boostable or trivial.

The learner we start from has the following property: there exists some $\delta_0 > 0$ such that
for every \(\epsilon>0\), given \(m_0(\delta_0, \epsilon)\) samples, it outputs a hypothesis \(h\) such that with probability at least \(1-\delta_0\),
\[
\bz(h)=\bigl(\mathrm{FP}(h),\mathrm{FN}(h)\bigr)\leq  \bz + \beps,
\]
for some $\bz\in Z$.
The procedure is based on centerpoints. We first recall the definition.
\begin{definition}[Centerpoint]
Let \(P=\{\bz_1,\dots,\bz_n\}\subseteq\mathbb{R}^2\) be a (multi)set of points,
where repetitions are allowed. A \emph{centerpoint} of \(P\) is a point
\(p\in\mathbb{R}^2\) such that every convex set
\(K\subseteq\mathbb{R}^2\) containing strictly more than \(2n/3\) points of \(P\)
(counting multiplicities) also contains \(p\).
\end{definition}
Observe that every centerpoint of \(\{\bz_1,\dots,\bz_n\}\) lies in
\(\conv(\{\bz_1,\dots,\bz_n\})\). The existence of centerpoints follows from the classical Centerpoint Theorem,
which goes back to~\cite{rado1946theorem} (see also
\cite{matouvsek2002lectures, pach2011combinatorial}).
\begin{theorem}[Centerpoint Theorem]\label{thm:centerpoint}
Every finite multiset \(P\subseteq\mathbb{R}^2\) admits a centerpoint.
\end{theorem}
In \(\mathbb{R}^2\), a centerpoint can be computed in polynomial time via
arrangements and halfspace depth; see, e.g.,
\cite{rousseeuw1998computing,matouvsek2002lectures}. Our  confidence-boosting procedure is given in \Cref{alg:centerpoint_confboost}.

\begin{algorithm}[ht]
\caption{Centerpoint-based confidence boosting}
\label{alg:centerpoint_confboost}
\begin{algorithmic}[1]\onehalfspacing
\REQUIRE oracle access to a learner \(\A\); parameters \(\delta,\epsilon>0\); base failure parameter \(\delta_0<1/3\)
\STATE Set \(n = C_{\delta_0}\log(1/\delta)\), for a sufficiently large constant \(C_{\delta_0}\)
\STATE Draw an independent validation sample \(V\) of size $N = C\frac{1}{\epsilon^2}\log\frac{n}{\delta}$,
for a sufficiently large universal numerical constant \(C\)
\FOR{$j=1,\ldots,n$}
    \STATE Draw an independent training sample \(S_j\) of size \(m_0(\epsilon/3, \delta_0)\)
    \STATE Run \(\A\) on \(S_j\) with accuracy parameter \(\epsilon/3\) to obtain \(h_j=\A(S_j)\)
    \STATE Estimate its error vector on \(V\):
    \[
    \widehat \bz_j
    =
    \bigl(\widehat{\mathrm{FP}}_V(h_j),\widehat{\mathrm{FN}}_V(h_j)\bigr)
    \]
\ENDFOR
\STATE Let \(\widehat \bz\) be a centerpoint of \(\{\widehat \bz_1,\dots,\widehat \bz_n\}\)
\STATE Find coefficients \(\lambda_1,\dots,\lambda_n\in \Delta_n\) such that $\widehat \bz=\sum_{j=1}^n \lambda_j\widehat \bz_j$.
\STATE Return the randomized hypothesis \(H\) which samples \(j\sim\lambda\)
and predicts using \(h_j\)
\end{algorithmic}
\end{algorithm}

We will now give our main result for confidence boosting, where the proof is deferred to \Cref{appendix:subsec:confboost}.
\begin{theorem}[Confidence boosting]
\label{thm:confidence_boosting}
Let \(Z\subseteq\mathbb{R}^2\) be convex, and let \(\A\) be a $(\delta_0,Z)$-learner for a class $\F$ with sample complexity $m_0$.
Then, for every \(\delta,\epsilon>0\),
\Cref{alg:centerpoint_confboost} requires  $n \cdot m_0(\epsilon/3, \delta_0) + N$ samples and
outputs a randomized hypothesis \(H\) such that,
with probability at least \(1-\delta\), there exists a point \(\bz\in Z\) for which
$
\bz(H) \leq \bz + \beps
$.
\end{theorem}

\paragraph{Discussion.}
The threshold \(\delta_0<1/3\) enters through the centerpoint argument: the
procedure needs more than a \(2/3\)-fraction of the empirical points to come from
successful calls to the learner. It would be interesting to understand whether
this is an inherent limitation of confidence boosting in this setting, or merely
an artifact of the centerpoint-based approach. In the classical \(0/1\)-loss
setting, confidence can be boosted from any constant failure probability below
\(1\), using validation. Here the difficulty is that the algorithm does not know
\(Z\), and therefore cannot directly validate whether a given call to the
\(Z\)-learner succeeded.

\paragraph{Use in the main boosting algorithm.}
In the sequel, we may assume that the \(Z\)-learner has arbitrarily small failure
probability. This is without loss of generality: before each invocation, we apply
\Cref{alg:centerpoint_confboost} with sufficiently small confidence parameter.
Thus, the main boosting algorithm can work with a learner whose sample size
corresponds only to constant confidence, while the confidence amplification is
handled modularly by the procedure above.

\section{Lower Bound}\label{sec:LB}
We next show that our sufficient condition for all \(Z\)-learners to be boostable, namely $\conv(\overline Z)\cap \C=\emptyset$, is also necessary. In particular, we prove the second item of \Cref{thm:main}, restated below.
\begin{theorem}\label{thm:LB}
If \(\conv(\overline Z)\cap \C\neq\emptyset,\) then there exists an algorithm \(\A\) that is a universal \(Z\)-learner (namely, a \(Z\)-learner for every \(\F\subseteq\Y^\X\)) and which outputs only random guesses.
\end{theorem}
The proof is deferred to \Cref{appendix:subsec:LB}, yet we give an overview of the proof next. The idea is to construct a learner that returns \emph{distribution-dependent} random guesses.  For every distribution \(\bq\in\Delta_\Y\), define
\[
C_{\bq}
=
\left\{
(z_1,z_2)\in[0,1]^2 :
\exists \bp\in\Delta_\Y
\text{ such that }
p_1q_2\le z_1
\text{ and }
p_2q_1\le z_2
\right\}.
\]
Thus, \(C_{\bq}\) is the set of guarantees achievable by random guessing against the label distribution \(\bq\). We then show in the proof that
$\C_{\bq}$ is in fact a halfspace.
Since $\conv(Z) \cap \C \ne \emptyset$, by the hyperplane separation theorem $\C_{\bq}$ does not separate $\conv(Z)$ and $\C$.
Since $\C \subseteq \C_{\bq}$, this implies $Z \cap \C_{\bq} \ne \emptyset$. Let then $(z_1,z_2) \in Z \cap \C_{\bq}$. Thus,  there exists $(p_+,p_-) \in \Delta_\Y$ such that
\begin{align}
  p_+ q_- \le z_1 \ \land \  p_- q_+ \le z_2\,.
\end{align}

Now let $\bq=(q_+,q_-)$ be the marginal distribution over $\Y$ induced by the concept $f$ and $\D$.
Consider the following learner $\A$.
First, $\A$ uses the sample to compute an estimate of $\bq$, which can easily get a total variation error of at most  $\epsilon$ with probability $1-\delta$ by using $\O(\epsilon^{-2} \log \delta^{-1})$ samples.
Then, $\A$ computes the distribution $\bp = (p_+,p_-) \in \Delta_\Y$ as above, using the estimate of $\bq$ in place of $\bq$, and returns the corresponding random guess $h_{\bp}$.
By construction $p_+ q_- \le z_1 + \epsilon$ and $p_- q_+ \le z_2 + \epsilon$, thus $\A$ is a $Z$-learner.
Moreover $\A$ returns only random guesses and is therefore not boostable.

\begin{ack}
MB, NCB, and EE acknowledge the support of the EU Horizon CL4-2022HUMAN-02 research and innovation action under grant agreement 101120237, project ELIAS (European Lighthouse of AI for Sustainability).
YM has received funding from the European Research Council (ERC) under the European Union’s Horizon 2020 research and innovation program (grant agreement No. 882396), by the Israel Science Foundation (1357/24),  the Blavatnik foundation at Tel Aviv University and a grant from the Tel Aviv University Center for AI and Data Science (TAD).
SM acknowledges support by Israel by ISF grant 2362/26, by the Technion Center for Machine Learning and Intelligent Systems (MLIS), and by the European Union (ERC, GENERALIZATION, 101039692). MT is supported by the German Research Foundation through the project 560788681.

\end{ack}

\bibliographystyle{plainnat}
\bibliography{bib}

@article{rado1946theorem,
  author = {Rado, Richard},
  title = {A theorem on general measure},
  journal = {Journal of the London Mathematical Society},
  volume = {21},
  number = {4},
  pages = {291--300},
  year = {1946}
}

@book{matouvsek2002lectures,
  author = {Matou{\v{s}}ek, Ji{\v{r}}{\'\i}},
  title = {Lectures on Discrete Geometry},
  publisher = {Springer},
  year = {2002}
}

@book{pach2011combinatorial,
  author = {Pach, J{\'a}nos and Agarwal, Pankaj K.},
  title = {Combinatorial Geometry},
  publisher = {Wiley},
  year = {1995}
}

@inproceedings{bressan2025dice,
  title={Of Dice and Games: A Theory of Generalized Boosting},
  author={Bressan, Marco and Brukhim, Nataly and Cesa-Bianchi, Nicol{\`o} and Esposito, Emmanuel and Mansour, Yishay and Moran, Shay and Thiessen, Maximilian},
  booktitle={The Thirty Eighth Annual Conference on Learning Theory},
  pages={596--640},
  year={2025},
  organization={PMLR}
}

@misc{schapire2012boosting,
  title={Boosting: Foundations and algorithms},
  author={Schapire, Robert E and Freund, Yoav},
  year={2012},
  publisher={MIT Press}
}

@article{Freund97decision,
	Author = {Yoav Freund and Robert E. Schapire},
	Bibsource = {dblp computer science bibliography, https://dblp.org},
	Journal = {J. Comput. Syst. Sci.},
	Number = {1},
	Pages = {119--139},
	Title = {A Decision-Theoretic Generalization of On-Line Learning and an Application to Boosting},
	Volume = {55},
	Year = {1997}}

@article{rousseeuw1998computing,
  author = {Rousseeuw, Peter J. and Ruts, Ida},
  title = {Computing the bivariate {T}ukey median},
  journal = {Statistica Sinica},
  volume = {8},
  number = {3},
  pages = {827--839},
  year = {1998}
}

@article{nikolaou2016cost,
	title={Cost-sensitive boosting algorithms: Do we really need them?},
	author={Nikolaou, Nikolaos and Edakunni, Narayanan and Kull, Meelis and Flach, Peter and Brown, Gavin},
	journal={Machine Learning},
	volume={104},
	number={2},
	pages={359--384},
	year={2016},
	publisher={Springer}
}

@article{sun2007cost,
	title={Cost-sensitive boosting for classification of imbalanced data},
	author={Sun, Yanmin and Kamel, Mohamed S and Wong, Andrew KC and Wang, Yang},
	journal={Pattern recognition},
	volume={40},
	number={12},
	pages={3358--3378},
	year={2007},
	publisher={Elsevier}
}

@article{landesa2012shedding,
	title={Shedding light on the asymmetric learning capability of {A}da{B}oost},
	author={Landesa-Vazquez, Iago and Alba-Castro, Jos{\'e} Luis},
	journal={Pattern Recognition Letters},
	volume={33},
	number={3},
	pages={247--255},
	year={2012},
	publisher={Elsevier}
}

@article{viola2001fast,
	title={Fast and robust classification using asymmetric {A}da{B}oost and a detector cascade},
	author={Viola, Paul and Jones, Michael},
	journal={Advances in neural information processing systems},
	volume={14},
	year={2001}
}

@article{cortes2020agnostic,
	title={Agnostic learning with multiple objectives},
	author={Cortes, Corinna and Mohri, Mehryar and Gonzalvo, Javier and Storcheus, Dmitry},
	journal={Advances in Neural Information Processing Systems},
	volume={33},
	pages={20485--20495},
	year={2020}
}

@book{jin2006multi,
	title={Multi-objective machine learning},
	author={Jin, Yaochu},
	volume={16},
	year={2006},
	publisher={Springer}
}

@article{kalai2012reliable,
	title={Reliable agnostic learning},
	author={Kalai, Adam Tauman and Kanade, Varun and Mansour, Yishay},
	journal={Journal of Computer and System Sciences},
	volume={78},
	number={5},
	pages={1481--1495},
	year={2012},
	publisher={Elsevier}
}

@article{kivinen1995learning,
	title={Learning reliably and with one-sided error},
	author={Kivinen, Jyrki},
	journal={Mathematical systems theory},
	volume={28},
	number={2},
	pages={141--172},
	year={1995},
	publisher={Springer}
}

@inproceedings{elkan2001foundations,
  author = {Charles Elkan},
  title = {The Foundations of Cost-Sensitive Learning},
  booktitle = {Proceedings of the 17th International Joint
               Conference on Artificial Intelligence},
  pages = {973--977},
  year = {2001}
}

@inproceedings{domingos1999metacost,
  author = {Pedro Domingos},
  title = {{MetaCost}: A General Method for Making
           Classifiers Cost-Sensitive},
  booktitle = {Proceedings of the Fifth ACM SIGKDD
               International Conference on Knowledge
               Discovery and Data Mining},
  pages = {155--164},
  year = {1999}
}

@inproceedings{zadrozny2003cost,
  author = {Bianca Zadrozny and John Langford
            and Naoki Abe},
  title = {Cost-Sensitive Learning by Cost-Proportionate
           Example Weighting},
  booktitle = {Proceedings of the Third IEEE
               International Conference on Data Mining},
  pages = {435--442},
  year = {2003}
}

@article{scott2012calibrated,
  author = {Clayton Scott},
  title = {Calibrated Asymmetric Surrogate Losses},
  journal = {Electronic Journal of Statistics},
  volume = {6},
  pages = {958--992},
  year = {2012}
}

@article{masnadi2011cost,
  author = {Hamed Masnadi-Shirazi and Nuno Vasconcelos},
  title = {Cost-Sensitive Boosting},
  journal = {IEEE Transactions on Pattern Analysis
             and Machine Intelligence},
  volume = {33},
  number = {2},
  pages = {294--309},
  year = {2011}
}

\newpage
\appendix
\section{Technical Claims}

 \subsection{Margin Equivalence} \label[appendix]{subsec:proof:margin_equivalence}
\begin{lemma}\label{lem:margin_equivalence} Let $Z,\C \subseteq [0,1]^2$ with $\C$ as defined in \cref{eq:coin_region_def}, and
$\operatorname{conv}(Z)\cap
\operatorname{int}_{[0,1]^2}(\mathcal C)
=\varnothing$.
Then,
    \[
    \min_{\substack{\bz'\in \conv(Z)\\ \bz\in \C}}
\|\bz-\bz'\|_\infty = \max_{\bw\in\Delta_2}\min_{\bz\in\conv(Z)}\left(w_1w_2 - \bw\cdot \bz\right) \;.
    \]
\end{lemma}
\begin{proof}

Let $\gamma'$ be defined as follows.
\begin{equation}
    \gamma' = \max_{\substack{\bw \in \Delta_2}} \left(\min_{\bz \in \C} (\bw \cdot \bz) - \max_{\bz \in \conv(Z)} (\bw \cdot \bz)\right) = \max_{\bw\in\Delta_2}\min_{\bz\in\conv(Z)}\left(w_1w_2 - \bw\cdot \bz\right),
\end{equation}
where the equality follows from \cref{eq:coin_region_def} via simple calculations.\footnote{The minimum is attained on the boundary and so denote $z_1=t^2$ and $z_2=(1-t)^2$. Then $\min_{t\in[0,1]} w_1t^2+w_2(1-t)^2$ is attained at $t=w_2$, giving value $w_1w_2^2+w_2w_1^2=w_1w_2$.}

By Claim~\ref{clm:sion} and the fact that
$\max_{\bw\in\Delta_2}\bw\cdot(\bz-\bz')
=\max_i(z_i-z'_i)$, we obtain
\[
\gamma'
=
\min_{\substack{\bz\in\mathcal C\\
\bz'\in\operatorname{conv}(Z)}}
\max_i(z_i-z'_i).
\]

Since $\mathcal C$ is coordinate-wise upward
closed, replacing $\bz$ by the coordinate-wise
maximum of $\bz$ and $\bz'$ shows that
\[
\min_{\substack{\bz\in\mathcal C\\
\bz'\in\operatorname{conv}(Z)}}
\|\bz-\bz'\|_\infty
=\max\{0,\gamma'\}.
\]
Finally, our assumption implies $\gamma'\ge0$:
otherwise, there would exist $\bz\in\mathcal C$
and $\bz'\in\operatorname{conv}(Z)$ with
$z'_i>z_i$ for both coordinates, implying
$\bz'\in\operatorname{int}_{[0,1]^2}(\mathcal C)$,
a contradiction.
\end{proof}

\begin{claim}\label{clm:sion}
Let \(\C,Z \subseteq \mathbb{R}^2\), such that \(\operatorname{conv}(Z)\) are both convex and compact. Then
\[
\max_{\bw\in \Delta_{\Y}}
\min_{\bz\in C,\ \bz'\in \operatorname{conv}(Z)}
\bw\cdot (\bz-\bz')
=
\min_{\bz\in \C,\ \bz'\in \operatorname{conv}(Z)}
\max_{\bw\in \Delta_{\Y}}
\bw\cdot (\bz-\bz') .
\]
\end{claim}

\begin{proof}
Let $K := C \times \operatorname{conv}(Z)$.
Since \(C\) and \(\conv(Z)\) are convex and compact, so is \(K\). Define $\ell:\Delta_{\Y}\times K \to \mathbb{R}$
by
\[
\ell(\bw,(\bz,\bz')) := \bw\cdot (\bz-\bz').
\]
Notice that it is bilinear.  Since \(\Delta_{\Y}\) and \(K\) are convex and compact, Sion's minimax theorem gives
\[
\max_{\bw\in \Delta_{\Y}}
\min_{(\bz,\bz')\in K}
\ell(\bw,(\bz,\bz'))
=
\min_{(\bz,\bz')\in K}
\max_{\bw\in \Delta_{\Y}}
\ell(\bw,(\bz,\bz')).
\]
Substituting the definitions of \(K\) and \(\ell\), the claim follows.
\end{proof}

\subsection{Missing Proofs of Section \ref{sec:boost}}\label[appendix]{appendix:missing_proofs:sec:boost}
This section contains the missing proofs and details which enable our main results from Section \ref{sec:boost}.

\begin{proof}[Proof of \Cref{lemma:zero_trainig_err}]
We will prove that, in some round $1\le r \le R$ of \Cref{alg:adaptive}, the invocation of \Cref{alg:binary_boost} never reaches line~\ref{step:fail}.
By \Cref{lem:binary_scalar_boosting} this implies that \Cref{alg:binary_boost} returns a hypothesis $\hat{h}_S$ consistent with $S$, in which case \Cref{alg:adaptive} reaches line~\ref{line:adaptive_return_h_S} and returns $\hat{h}_S$ itself, yielding the claim.

Since $\widehat m =
m_0(\gamma/3,\delta/(2TR))$, a union bound
implies that, with probability at least
$1-\delta/2$, every call to $\mathcal A$
returns a hypothesis $h$ whose actual
error vector $\bz=\operatorname{errs}_{D_t,f}(h)$
satisfies $\bz\le \bar \bz+\gamma/3$
for some $\bar \bz\in Z$. We condition on this event for the remainder
of the proof.

Since $\bw\in\Delta_2$, the definition of the
validity function gives
\begin{equation}
    v(w_1;\bz)\ge v(w_1;\bar \bz)-\gamma/3.
\end{equation}
In particular, by the definition of $w_1^*$
and the weak learner's accuracy guarantee,
\begin{equation}\label{eq11}
    v(\bw_1^*;\bz)
\ge v(w_1^*;\bar \bz)-\gamma/3
\ge \gamma^*-\gamma/3
\ge 5\gamma/3,
\end{equation}
where the last inequality uses
$\gamma\le\gamma^*/2$.

Suppose that at round $r \in \{1,\ldots,R\}$ \Cref{alg:adaptive} reaches line~\ref{step:witness}.
Then, by construction of \Cref{alg:binary_boost}, it receives $\bz$ such that $v(w_1^{(r)};\bz) < \gamma$.
On the other hand we have shown that $v(w_1^*;\bz) \ge 5\gamma/3$.
Using these facts, we claim that $w_1^*$ belongs to the updated interval $[L_{r+1},U_{r+1}]$, see the lines after line~\ref{step:witness}.
First, note that the function $v(\cdot \,;\bz) : \RR \to \RR$ is a quadratic polynomial, and
its derivative at $w_1^{(r)}$ is:
\[
\Delta := \left.\frac{\partial}{\partial w}v(w;\bz)\right|_{w=w_1^{(r)}}
= 1-2w_1^{(r)}-z_1+z_2 .
\]
If \(\Delta>0\), then \(v(\cdot\,,\bz)\) is increasing at \(w_1^{(r)}\); it follows that \(w_1< w_1^*\). Symmetrically, if $\Delta < 0$ then \(v(\cdot;\bz)\) is decreasing at \(w_1^{(r)}\) and thus \(w_1^*<w_1\).
This shows that $w_1^* \in [L_{r+1},U_{r+1}]$, as claimed.

By construction, $U_r-L_r=2^{1-r}$.
Since $w_1^*\in[L_r,U_r]$ at every round
and $w_1^{(r)}$ is the midpoint,
our choice of $R$ ensures that
\[
|w_1^{(R)}-w_1^*|
\le 2^{-R}\le\gamma/3.
\]

Now let $\bz$ be any actual error vector
returned by a successful call to the
weak learner.    Note that $v(\cdot;\bz)$ is $2$-Lipschitz
for every $\bz\in[0,1]^2$, and so we obtain
\[
\begin{aligned}
v(w_1^{(R)};\bz)
&\ge v(w_1^*;\bz)-2|w_1^{(R)}-w_1^*| \ge \frac{5\gamma}{3}
           -\frac{2\gamma}{3} =\gamma,
\end{aligned}
\]
where the second inequality follows by \eqref{eq11} and   $|w_1^{(R)}-w_1^*|\le\gamma/3$.

Thus, at round $R$, every successful call
to the weak learner produces an error
vector satisfying the validity test in
\Cref{alg:binary_boost}. Consequently, \Cref{alg:binary_boost}
cannot reach its failure step and must
return a hypothesis consistent with $S$.
This completes the proof.
\end{proof}

\begin{proof}[Proof of \Cref{thm:adaptive_boost_formal}]
First, we apply \Cref{lemma:zero_trainig_err} to  obtain that, with probability at least $1-\delta/2$, the hypothesis $\widehat h_S$ is consistent with the labeled sample $S$.
Next, we apply a standard compression-based generalization argument (see, e.g., Theorem~2.8 from \citet{schapire2012boosting}).  To this end, note that one can construct a compression scheme for $\widehat h_S$ of size $\kappa$ equal to the total size of the samples on which $\A$ is invoked, that is, $\kappa = \widehat m \cdot TR$, where $\widehat{m} = m_0\left(\tfrac\gamma3,\tfrac{\delta}{2TR}\right)$.
By Theorem~2.8 from \citet{schapire2012boosting} with $\delta/2$ in place of $\delta$ we get that, with probability at least $1-\delta/2$ any consistent hypothesis $h$ produced by a $\kappa$-size compression scheme satisfies,
\begin{align}
    \Pr_{x \sim \D}\Big[h(x) \neq f(x) \Big] \le \frac{\kappa \ln(m) + \ln(2/\delta)}{m-\kappa} \;.
\end{align}
By union bound we overall get that with probability at least $1-\delta$, the above inequality holds for the hypothesis $\widehat h_S$ produced by \Cref{alg:adaptive}.
Straightforward calculations show that the right-hand side is at most $\eps$ whenever:
\begin{align}
    m &\ge \frac{\ln(2/\delta)}{\eps} + \kappa \cdot \left(1+\frac{\ln(m)}{\eps}\right) = \frac{\ln(2/\delta)}{\eps} + m_0\left(\tfrac{\delta}{2TR}\right) \cdot TR \cdot \left(1+\frac{\ln(m)}{\eps}\right) \;.
\end{align}
We use the standard inequality \(\ln m \le \alpha m+\ln(1/\alpha)-1\),  for any \(\alpha>0\), to get that the requirement \(m\ge A+B\ln m\) is satisfied whenever \(m\ge 2A+2B\ln(2B)\), by plugging \(\alpha=\tfrac{1}{2B}\) in the above.
It therefore suffices to take
\[
m \ge
\frac{4}{\epsilon}
\left(
\ln\frac{2}{\delta}
+
m_0\!\left(\tfrac{\delta}{2TR}\right)TR
\cdot
\ln\left(
\frac{4m_0\!\left(\frac{\delta}{2TR}\right)TR}{\epsilon}
\right)
\right).
\]
The bound on $m$  then follows by plugging in the choice of $T$ and $R$. It suffices to take
\[
m \ge
\frac{8}{\epsilon}\ln\frac{2}{\delta}
+
\frac{512R}{\epsilon\gamma^2}
m_0\!\left(\frac{\delta}{2TR}\right)
\ln^2\left(
\frac{
512R m_0\!\left(\frac{\delta}{2TR}\right)
}{
\epsilon\gamma^2
}
\right) = \Tilde{O}\left(\frac{m_0(\tfrac{\delta}{2TR})}{\epsilon\gamma^2}\right).
\]
\end{proof}

\subsection{Missing Proofs of \Cref{sec:confboost}}\label[appendix]{appendix:subsec:confboost}
In what follows we will prove \Cref{thm:confidence_boosting}, showing that the total sample size used by the procedure is
\[
n \cdot m_0(\delta_0,\epsilon/3) + N
=
O_{\delta_0}\!\left(m\left(\frac{\epsilon}{3}\right)\log\left(\frac{1}{\delta}\right)\right)
+
O\!\left(
\frac{1}{\epsilon^2}\log\frac{\log(1/\delta)}{\delta}
\right).
\]
where the dependence on \(\delta_0\) arises from the choice
\[
n = C_{\delta_0}\log(1/\delta), \qquad
C_{\delta_0} = \Theta\!\left(\frac{1}{(1/3-\delta_0)^2}\right),
\]
as follows from a Chernoff bound.
\begin{proof}[Proof of \Cref{thm:confidence_boosting}]
Let \(h_1,\dots,h_n\) be the hypotheses produced by the \(n\) independent calls to \(\A\), and let
\[
\bz_j=\bz(h_j).
\]
In the next definition we use the following standard notation: for sets \(A,B\subseteq\mathbb{R}^2\), the \emph{Minkowski sum} is defined as
\(A+B \triangleq \{a+b : a\in A,\, b\in B\}.\)

Call an index \(j\) good if
\[
\bz_j \in Z + [-\epsilon/3,+\epsilon/3]^2,
\]
By assumption, each index is good with probability at least \(1-\delta_0\), independently. Since \(\delta_0<1/3\), by a Chernoff bound, for
\[
n = C_{\delta_0}\log(1/\delta),
\]
with
\[
C_{\delta_0} = \Theta\!\left(\frac{1}{(1/3-\delta_0)^2}\right),
\]
it follows that with probability at least \(1-\delta/2\), more than \(2n/3\) of the indices are good.

Next, consider the empirical estimates \(\widehat \bz_j\) computed on the
validation sample \(V\). Since \(V\) is independent of the hypotheses
\(h_1,\dots,h_n\), standard concentration and a union bound imply that, since
\[
N =
C\frac{1}{\epsilon^2}\log\frac{n}{\delta},
\]
with \(C\) sufficiently large numerical constant,
then with probability at least \(1-\delta/2\), for
every \(j\in[n]\),
\[
\|\widehat \bz_j-\bz_j\|_\infty\le \epsilon/3.
\]

Assume that both high-probability events hold (which happens with probability at least $1-\delta$). Then, for every good index \(j\),
\[
\widehat \bz_j \in Z + [-2\epsilon/3,+2\epsilon/3]^2.
\]
Thus, more than \(2n/3\) of the empirical estimates lie in the convex set (recall that w.l.g.\ $Z$ is convex)
\[
Z + [-2\epsilon/3,+2\epsilon/3]^2.
\]
By the definition of a centerpoint,
\[
\widehat \bz\in Z + [-2\epsilon/3,+2\epsilon/3]^2.
\]
Since \(\widehat \bz\in \conv(\{\widehat \bz_1,\dots,\widehat \bz_n\})\), the coefficients \(\lambda\) exist, and the output hypothesis satisfies
\[
\bz(H)=\sum_{j=1}^n \lambda_j \bz_j,
\qquad
\widehat \bz=\sum_{j=1}^n \lambda_j \widehat \bz_j.
\]
Therefore,
\[
\|\bz(H)-\widehat \bz\|_\infty
\le
\sum_{j=1}^n \lambda_j\|\bz_j-\widehat \bz_j\|_\infty
\le \epsilon/3.
\]
Combining,
\[
\bz(H)\in Z + [-\epsilon,+\epsilon]^2.
\]
The total failure probability is at most \(\delta\), completing the proof.
\end{proof}

 \subsection{Missing Proofs of \Cref{sec:LB}}\label[appendix]{appendix:subsec:LB}

In this subsection we give the missing proofs of \Cref{sec:LB}, concerning the lower bound of our main result.

\begin{proof}[Proof of \Cref{thm:LB}]
The idea is to construct a learner that returns \emph{distribution-dependent} random guesses. For every distribution \(\bq\in\Delta_\Y\), define
\[
C_{\bq}
=
\left\{
(z_1,z_2)\in[0,1]^2 :
\exists \bp\in\Delta_\Y
\text{ such that }
p_1q_2\le z_1
\text{ and }
p_2q_1\le z_2
\right\}.
\]
Thus, \(C_{\bq}\) is the set of guarantees achievable by random guessing against the label distribution \(\bq\).

The explicit characterization of \(\C\) plays a crucial role here.
By \Cref{clm:C_q_def} (stated and proven below),
\[
C_{\bq}
=
\left\{
\mathbf e\in[0,1]^2 :
q_1e_1+q_2e_2\ge q_1q_2
\right\},
\]
and hence \(C_{\bq}\) is the intersection of $[0,1]^2$ with a halfspace.
Therefore, $[0,1]^2\setminus C_{\bq}$ is also convex.

Recall that
\(
\C=
\left\{
(e_1,e_2)\in[0,1]^2 :
\sqrt{e_1}+\sqrt{e_2}\ge1
\right\}.
\)
A direct calculation shows that every \((e_1,e_2)\in\C\) satisfies
\[
q_1e_1+q_2e_2\ge q_1q_2,
\]
and therefore
\(
\C\subseteq C_{\bq}
\)
for every $\bq\in\Delta_\Y$.
Now, assume
\(\conv(\overline Z)\cap\C\neq\emptyset.\)
Since \(\C\subseteq C_{\bq}\), it follows that
\(
\conv(\overline Z)\cap C_{\bq}\neq\emptyset.
\)
We claim that therefore
\[
\overline Z\cap C_{\bq}\neq\emptyset.
\]
Indeed, otherwise \(\overline Z\subseteq [0,1]^2\setminus C_{\bq}\), and since $[0,1]^2\setminus C_{\bq}$ is also convex, we have
\[
\conv(\overline Z)\subseteq [0,1]^2\setminus C_{\bq},
\]
contradicting \(\conv(\overline Z)\cap C_{\bq}\neq\emptyset.\)

We conclude that there exists
\((z_1,z_2)\in\overline Z\)
and \(\bp=(p_1,p_2)\in\Delta_\Y\) such that
\begin{equation}\label{eq:LB_pqz}
p_1q_2\le z_1,
\qquad
p_2q_1\le z_2.
\end{equation}

We now construct the learner.
Let \(\bq=(q_1,q_2)\) denote the distribution over labels induced by \(f\) and~\(\D\): that is,
$q_1 = \Pr_{x\sim \D}[f(x)=1]$ and $q_2 = \Pr_{x\sim \D}[f(x)=-1]$. Given a sample, the learner first computes an estimate \(\hat\bq\in\Delta_\Y\) such that
\[
\Pr\bigl(\|\hat\bq-\bq\|_\infty\le\epsilon/2\bigr)\ge1-\delta,
\]
using \(O(\epsilon^{-2}\log(\delta^{-1}))\) samples. Next, applying the above argument to \(\hat\bq\), the learner chooses some
\[
\hat z=(\hat z_1,\hat z_2)\in\overline Z\cap C_{\hat\bq}
\]
together with a distribution \(\hat\bp\in\Delta_\Y\) satisfying
\[
\hat p_1\hat q_2\le\hat z_1,
\qquad
\hat p_2\hat q_1\le\hat z_2.
\]
Finally, \(\A\) outputs an hypothesis which is a random label according to \(\hat\bp\).

Assume now that
\(
\|\hat\bq-\bq\|_\infty\le\epsilon/2
\)
(which happens with probability at least $1-\delta$).
Then
\[
\hat p_1q_2\le \hat z_1+\epsilon/2,
\qquad
\hat p_2q_1\le \hat z_2+\epsilon/2.
\]
Since \(\hat z\in\overline Z\), for every \(\eta>0\) there exists
\(
\bz'=(z_1',z_2')\in Z
\)
such that
\(
\|\bz'-\hat \bz\|_\infty\le\eta.
\)
Therefore,
\[
\hat p_1q_2\le z_1'+\epsilon/2+\eta,
\qquad
\hat p_2q_1\le z_2'+\epsilon/2+\eta.
\]
For \(\eta= \epsilon/2\)  it follows that \(\A\) is a \(Z\)-learner.
\end{proof}

\begin{claim}\label{clm:C_q_def}
For every \(\bq \in\Delta_2\) the following sets are identical:
\begin{align*}
C_{\bq}
&=
\left\{
\mathbf e \in [0,1]^2 :
\exists \bw \in\Delta_2 :
w_1q_2\le e_1
\wedge
w_2q_1\le e_2
\right\},
\\
C_{\bq}'
&=
\left\{
\mathbf e\in [0,1]^2 :
q_1e_1+q_2e_2\ge q_1q_2
\right\}.
\end{align*}
\end{claim}

\begin{proof}
If \(q_1=0\) or \(q_2=0\), then
\[
C_{\bq}=C_{\bq}’=[0,1]^2.
\]
Otherwise, \(\mathbf e \in C_{\bq}\) iff there exists \(t\in(0,1)\) such that
\[
tq_2\le e_1
\qquad\text{and}\qquad
(1-t)q_1\le e_2.
\]
Equivalently,
\[
t\le \frac{e_1}{q_2}
\qquad\text{and}\qquad
t\ge 1-\frac{e_2}{q_1}.
\]
Such a \(t\) exists iff
\[
1-\frac{e_2}{q_1}\le \frac{e_1}{q_2},
\]
which is equivalent to
\[
q_1e_1+q_2e_2\ge q_1q_2.
\]
\end{proof}

\end{document}